\documentclass[runningheads]{llncs}

\usepackage{eccv}

\usepackage{eccvabbrv}

\usepackage{graphicx}
\usepackage{booktabs}
\usepackage{multirow}
\usepackage{makecell}

\usepackage[accsupp]{axessibility}  % Improves PDF readability for those with disabilities.

\usepackage[pagebackref,breaklinks,colorlinks,citecolor=eccvblue]{hyperref}
\usepackage{orcidlink}

\begin{document}

\title{FipTR: A Simple yet Effective Transformer Framework for Future Instance Prediction in Autonomous Driving} 

\titlerunning{FipTR for Future Instance Prediction in Autonomous Driving}

\author{Xingtai Gui\inst{1} \and
Tengteng Huang \inst{1} \and
Haonan Shao \inst{1} \and
Haotian Yao \inst{1} \and
Chi Zhang \inst{1}
}

\authorrunning{X.~Gui et al.}

\institute{Mach-Drive}

\maketitle

% abstract 
\begin{abstract}
  The future instance prediction from a Bird's Eye View(BEV) perspective is a vital component in autonomous driving, which involves future instance segmentation and instance motion prediction. Existing methods usually rely on a redundant and complex pipeline which requires multiple auxiliary outputs and post-processing procedures. Moreover, estimated errors on each of the auxiliary predictions will lead to degradation of the prediction performance. In this paper, we propose a simple yet effective fully end-to-end framework named Future Instance Prediction Transformer(FipTR),  which views the task as BEV instance segmentation and prediction for future frames. We propose to adopt instance queries representing specific traffic participants to directly estimate the corresponding future occupied masks, and thus get rid of complex post-processing procedures. Besides, we devise a flow-aware BEV predictor for future BEV feature prediction composed of a flow-aware deformable attention that takes backward flow guiding the offset sampling. A novel future instance matching strategy is also proposed to further improve the temporal coherence. Extensive experiments demonstrate the superiority of FipTR and its effectiveness under different temporal BEV encoders. The code is available at \href{https://github.com/TabGuigui/FipTR}{https://github.com/TabGuigui/FipTR}.
\end{abstract}

\section{Introduction}
\label{sec:intro}

Future instance prediction from a Bird's Eye View (BEV) is a challenging and vital task for autonomous driving. It aims at predicting the occupied area and potential motion state of interested road participants around the ego car in the future, given the observed surrounding images of several history frames. Future instance prediction provides enriched and valuable information of the dynamic environment with the potential occupied situation, thus playing a key role for the downstream planning strategies.

Inspired by the success of BEV perception paradigm in object detection~\cite{bevdet,bevformer} and map segmentation~\cite{bevsegformer, baeformer}, many approaches adapt the BEV architecture to the future instance prediction task~\cite{fiery,beverse,powerbev,tbPformer}. FIERY~\cite{fiery} and BEVerse~\cite{beverse} utilize RNN to explicitly predict future BEV representation based on the last BEV map in an iterative manner. In contrast, PowerBEV~\cite{powerbev} and TBPFormer~\cite{tbPformer} employ strong backbone networks to extract multi-scale features from history frames, and directly output multi-frame BEV predictions in parallel, leaving the temporal relation between adjacent frames implicitly modeled. Although these methods demonstrate promising performance, they usually involve multiple auxiliary predictions and complex post-processing procedures. Concretely, they first estimate instance centerness and offsets, and then use clustering algorithms to group the BEV grids into instances. Moreover, to correlate a unique instance that appeared in multiple frames, they predict a motion flow and adopt post-processing to assign the same ID to the identical instance across frames in an offline manner. For such a complex pipeline with multiple post-processing procedures, the estimated error on each of the auxiliary predictions will lead to degradation of the prediction performance. Therefore, exploring a simple and effective framework is intuitive and desirable.

In this paper, we propose a simple yet effective fully end-to-end framework named \textbf{F}uture \textbf{I}nstance \textbf{P}rediction \textbf{T}ransforme\textbf{R}(FipTR). Inspired by the pioneering end-to-end detection framework~\cite{detr}, FipTR adopts a group of instance queries to directly estimate the future occupied masks and motion state for interested instances in the BEV plane, without the necessity of centerness estimation and clustering process~\cite{fiery,beverse,powerbev,tbPformer}. In this way,
FipTR gets rid of redundant outputs and complex post-processing procedures and thus shares an elegant and fully end-to-end pipeline. 

Based on the end-to-end design, we propose two important strategies, \textit{i.e.}, flow-aware BEV predictor, and a future instance matching mechanism for further improvement on the temporal coherence of prediction results for a unique instance appearing in multiple future frames. The flow-aware BEV predictor is responsible for updating the current BEV map, via establishing the spatial correspondency between the previous BEV map and the current one. To this end, we bring the idea of backward flow into BEV perception, which provides the offsets along the x- and y-axis from a specific grid in the BEV map to its corresponding grid in the previous BEV map. In this way, we can establish a dense mapping between grids from two adjacent BEV maps. The flow-aware BEV predictor is composed of two core parts, one is the flow prediction module for estimating the backward flow, and the other is flow-aware deformable attention. We use the backward flow to guide the offset prediction in deformable attention and ensure that grids occupied by an identical instance from different frames are closely related, which is verified valuable for temporally coherent and interpretable prediction.

Different from common Hungarian matching based on the box or single mask cost~\cite{detr, maskformer, vistr}, future instance matching is designed to always assign an identical ground truth object across future frames to the same instance query by attaching multi-frame mask cost, which naturally enhances the temporal coherence across adjacent frames. Compared with independently applying Hungarian matching to each frame, the proposed method alleviates the risk of assigning a ground truth object to multiple instance queries, and vice versa, leading to more stable matching and more precise prediction results.

To summarize, the main contributions of this work are:

\begin{itemize}

    \item We propose a fully end-to-end future instance prediction framework named FipTR, which adopts instance query to directly estimate the occupied masks eliminating the need for complex post-processing procedures.
    
    \item We propose a novel flow-aware BEV predictor module, which generates a more temporally coherent BEV map by flow-aware deformable attention guided by an estimated backward flow from the current BEV map to the previous one. 

    \item We design a future instance matching strategy to assign an object appearing in multiple frames to a unique instance query, which naturally improves temporal consistency.
    
    \item Extensive experiments demonstrate the superiority of FipTR and its effectiveness under different BEV encoders.
    
\end{itemize}

%------------------------------------------------------------------------
\section{Related Work}

\subsection{BEV Perception}
BEV Perception is becoming a promising task in autonomous driving, especially the vision-based formula for its low-price and unified characteristics. The core challenge is to construct a mapping from the surrounding cameras 2D features to 3D features considering the multi-view feature fusion and depth estimation. LSS-based methods~\cite{lss, bevdet, bevdepth}, achieve satisfactory performance by lifting the 2D features with accurate depth estimation. Transformer-based methods~\cite{bevformer, bevformerv2, crosstransformer} focus on the interactions between different feature spaces through the attention mechanism to obtain effective BEV representations. Considering the temporal information fusion, BEVFormer~\cite{bevformer} proposes a temporal cross-attention mechanism to achieve multi-frame feature interaction, while BEVDet4D~\cite{bevdet4d} leverages feature warping with egopose to achieve multi-frame alignment.

\subsection{Segmentation Transformer}
The transformer-based image segmentation methods~\cite{maskformer, mask2former, oneformer, knet} utilize the transformer encoder-decoder structure to achieve promising performance. Considering the success of image segmentation with attention mechanism, some video instance segmentation works~\cite{vistr, seqformer, genvis} incorporate the idea of object query into video streams for end-to-end formula. VisTR~\cite{vistr} is the first work with transformer that assigns a query to each instance and designs an instance sequence matching scheme to realize end-to-end instance segmentation. SeqFormer~\cite{seqformer} further incorporates the detection branch for better performance. Even though GenVIS~\cite{genvis} introduces a multi-mask matching cost for temporal consistency, the mask output is computed with observed frames, and the effectiveness of the multi-mask matching for future frames is still unexplored. In the field of autonomous driving, transformer-based semantic map segmentation methods also achieve impressive success~\cite{bevsegformer, baeformer, lara, unifusion}. However, these works mainly focus on the perception at the current moment while future instance prediction is still overlooked which is important for understanding the surrounding environment in the future.

\subsection{Future Instance Prediction}
Predicting the future surrounding environment from a BEV perspective is an essential part of autonomous driving. Compared to agent-based trajectory prediction~\cite{intentnet, pnpnet, densetnt}, the agent-free formula focuses on future grid occupancy to better understand the dynamic environment. Recently, BEV future instance prediction based on cameras has received increasing attention, which requires the model to simultaneously predict future grid occupancy and instance segmentation. FIERY~\cite{fiery} firstly defines this task and proposes to fuse multi-frame camera semantic information with RNN module to predict the future state. BEVerse~\cite{beverse} proposes a multi-task framework that simultaneously performs detection, mapping, and motion prediction with a more effective prediction scheme based on RNN. TBPFormer~\cite{tbPformer} suggests a novel BEV temporal encoder directly mapping multi-frame images to the same BEV spatial-temporal space and a prediction module based on transformer that markedly enhances the prediction performance. Although these works are effective in predicting future occupancy states, it is indispensable to predict a large number of auxiliary results, such as offset, centerness, and motion flow to achieve precise ID assignment and the continuity of instance IDs between frames with complex post-processing. The end-to-end UniAD~\cite{uniad} framework also considers future instance prediction as a part of the entire pipeline and proposes a transformer-based solution that can predict instance segmentation without post-processing. However, UniAD heavily depends on the first stage to obtain effective track queries and lacks an in-depth discussion on this BEV feature prediction. To the best of our knowledge, our work is the first fully end-to-end framework in BEV future instance prediction. 

\section{Methodology}

\begin{figure*}[ht]
  \centering
   \includegraphics[width=1.0\linewidth]{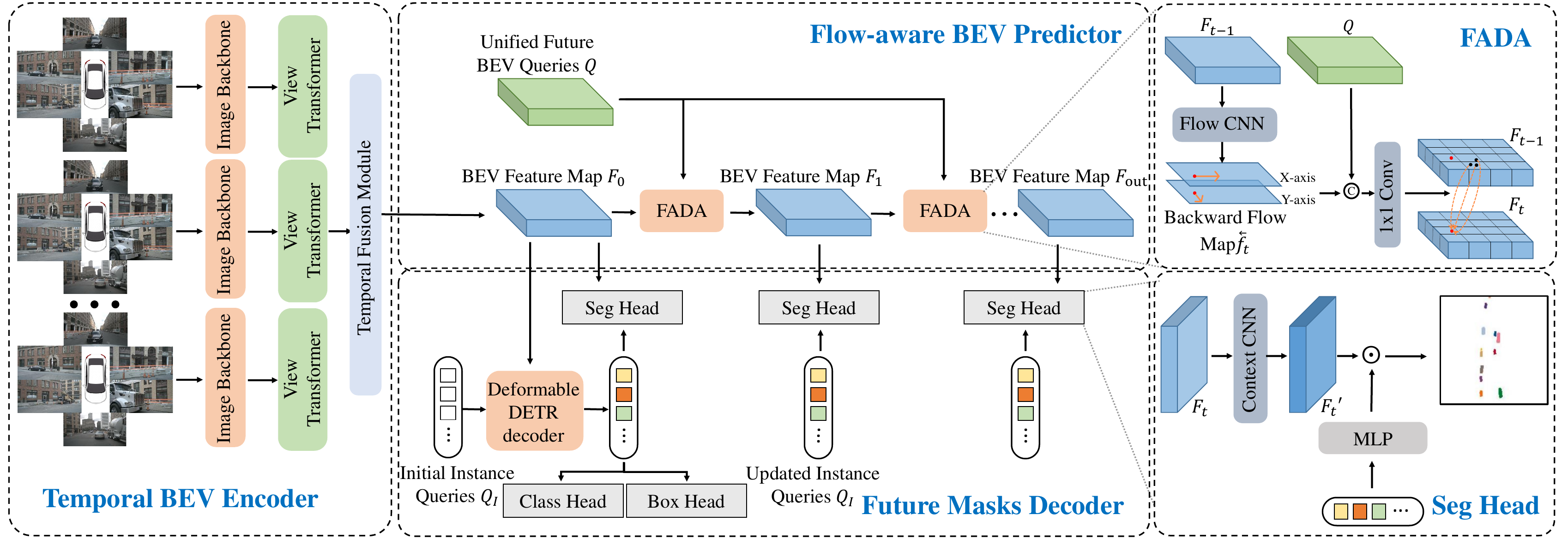}

   \caption{The overall architecture of FipTR. Given the multi-frame surrounding images, a temporal BEV Encoder including image backbone, view transformer and temporal fusion module, generates the current frame BEV feature map $F_0$. The flow-aware BEV predictor takes $F_0$ and unified future BEV query as input and updates the future BEV feature map iteratively. The flow-aware deformable attention(FADA) takes the predicted backward flow into consideration and generates motion-aware sampling offsets for BEV prediction. The future masks decoder outputs the future instance segmentation by conducting element-wise product between predicted BEV feature sequence and instance queries and predicting the corresponding category and 3D box attributes.}
   \label{fig:overall}
\end{figure*}

%-------------------------------------------------------------------------
\subsection{Overview Architecture}

Future instance prediction takes as input a group of multi-view images $\mathrm{I} \in \mathbb{R}^{T_{in} \times N \times 3 \times H \times W } $ from $T_{in}$ consecutive frames with N different view images per frame, and predicts dense BEV instance segmentation masks for the following $T_{out}$ future frames. The predicted masks indicate the occupied areas and potential motion of road participants around the ego car, which provides valuable information for planning.

In this section, we introduce the proposed simple yet effective framework named FipTR, which predicts the future segmentation masks in a fully end-to-end manner. \cref{fig:overall} illustrates the overall architecture of FipTR, consisting of three main parts, namely Temporal BEV Encoder, Flow-aware BEV Predictor, and Instance Masks Decoder. Moreover, we propose a novel matching strategy, Future Instance Matching, to further ensure temporal coherence across future frames.

\subsection{Temporal BEV Encoder}

The Temporal BEV Encoder is composed of three basic parts: an image backbone~(\textit{e.g.}, ResNet~\cite{resnet}, SwinTransformer~\cite{swin}), a view transformer, and a temporal fusion module. The image backbone takes images $\mathrm{I}$ as input and extracts their semantic feature maps, which are then converted into BEV feature maps by the following view transformer. To fully make use of the temporal cues, a temporal fusion module is applied to compress the BEV feature maps across consecutive frames into a single BEV feature map $F_0 \in \mathbb{R}^ {C \times H \times W}$.

We provide two variants for different temporal BEV encoders in FipTR based on BEVDet4D~\cite{bevdet4d} and BEVFormer~\cite{bevformer} respectively, which are two popular temporal BEV perception paradigms. BEVDet4D first generates BEV feature maps via view transformer, LSS~\cite{lss}, and then combines coordinate-aligned history BEV maps in a concatenation manner, while BEVFormer adopts cross attention mechanism as the view transformer and temporal fusion module respectively. Since the temporal BEV encoder is not the core contribution of this paper, we refer the reader to their original papers for more details.

\subsection{Flow-aware BEV Predictor}

The BEV feature map $F_0$ encodes the rich temporal-fused semantic features of $T_{in}$ history frames and indicates the observed motion states of interested objects around the ego-car, as well as their potential positions in the future. To achieve future predictions, it is crucial to obtain effective BEV feature maps at future frames. Motivated by this, we propose a flow-aware BEV predictor, which takes $F_0$ as the initial BEV feature, to update the BEV feature map iteratively for the following $T_{out}$ future frames. For the convenience of description, we denote $F_t$ as the BEV feature at timestamp $t$ in the future, where $t \in \{1, 2, ..., T_{out} \}$. The flow-aware BEV predictor mainly consists of three components, namely unified future BEV query, flow prediction module, and flow-aware deformable attention.

\noindent\textbf{Unified future BEV queries.}~We predefined a group of grid-shaped learnable parameters $Q\in \mathrm{R}^{C \times H \times W }$ which represent the initial BEV feature for a future frame. We denote $Q_{t,p}$ as the BEV query located in a specific position $p=\{x,y\}$ at timestamp $t$. We share $Q$ across frames, \textit{i.e.}, $Q_{1}=Q_{2}=...=Q_{T_{out}}$. Following common practices~\cite{bevformer}, we add $Q_{t}$ with a learnable positional encoding, before feeding them into the following flow-aware deformable attention module to generate the updated BEV feature $F_{t}$.

\noindent\textbf{Flow prediction module.}~Given the BEV feature map from the previous frame $F_{t-1}$, we use a flow prediction module to produce the backward flow map $\overleftarrow{f_{t}}$ which indicates the spatial correspondence relationship between the grids of $F_t$ and those of $F_{t_1}$~\cite{occflow}. Concretely, $\overleftarrow{f_{t}} \in \mathbb{R}^{2 \times H \times W}$ shares the same height $H$ and width $W$ with BEV feature map, while its two channels represent the offsets along the X- and Y-axis respectively.  Take a specific grid with the coordinate $(x, y)$ in $F_t$ as an example, its corresponding grid in $F_{t-1}$ is the one with the coordinate $(x+\Delta x, y+\Delta y)$ and $(\Delta x, \Delta y)$ is the backward flow. In a similar way, we can establish the dense spatial correspondence relationship $\overleftarrow{f_{t}}$, which will be used to guide the offset sampling from $F_{t-1}$ to $F_{t}$. The predicted flow map is supervised by the ground truth backward flow.
% Check

Thanks to the enriched temporal information encoded by the temporal BEV encoder, we empirically find that adopting a light and simple subnetwork composed of two standard residual CNN blocks is enough to produce a valuable backward flow map. It should be noticed that we use the same yet not weight-shared subnetworks for predicting the backward flow at different timestamps $t$, which is more adaptive and improves the performance. 

\noindent\textbf{Flow-aware deformable attention.} We design the novel Flow-aware Deformable Attention(FADA) module to transfer BEV feature from the previous frame $F_{t-1}$ into $F_{t}$. To this end, we leverage the predicted backward flow $\overleftarrow{f_{t}}$ which provides information-rich cues of correspondence relations between grids from these frames, for better guidance of the transformation procedure.

Before delving into the details of FADA, we first revisit the vanilla deformable attention(DA) mechanism, which takes $Q_t$ paired with its reference point $p$ and $F_{t-1}$ as inputs, and outputs $F_{t}$. The calculation process can be formulated as:

\begin{equation}
    \label{eq: dattn}
    \mathrm{DA}(Q_{t,p}, p, F_{t-1}) = \\
    \sum_{m=1}^M W_m \sum_{k = 1}^K A_{mk} W'_mF_{t-1}(p+ \Delta p_{mk})
\end{equation}

where $p$ is a reference point and $Q_{t,p}$ is the corresponding query, respectively. $m$ indexes the attention head, and $M$ denotes the number of attention heads. $k$ indexes the sampled key and $K$ is the number of sampled keys for each head. $W_m \in \mathbb{R}^{C \times (C / M)}$ and  $W_m' \in \mathbb{R}^{(C/M) \times C}$ are learnable weights. $A_{mk}$ is the attention weight obtained via linear projection over the query and normalized by $\sum_{k = 1}^K A_{mk} = 1$. $\Delta p_{mk}$ is the sampling offsets for a specific reference point, and $F_{t-1}(p+ \Delta p_{mk})$ refers to the input BEV feature at location $p+ \Delta p_{mk}$.

In the original deformable attention, $\Delta p_{mk}$ is generated as follows:

\begin{equation}
    \Delta p_{mk} = W \cdot Q_p,
\end{equation}

where $W$ is a learnable parameter. While for FADA, we take the predicted backward flow $\overleftarrow{f_{t}}$ into consideration to achieve more precise offset sampling for the transformation from $F_{t-1}$ to $F_t$. Hence, the sampling offsets can be formulated as follows:

\begin{equation}
    \Delta p'_{mk} = W \cdot \mathrm{concat}(\overleftarrow{f_{t}}, Q_p),
\end{equation}

In this way, FADA explicitly leverages the cues of motion states embedded in the history BEV features and thus brings more interpretability for the prediction process.

\subsection{Instance Masks Decoder}

Instance Masks Decoder takes the BEV feature map $F_t$ in all frames and fixed-number instance queries as input. The instance queries are a set of learnable embeddings, $\mathrm{Q}_I \in \mathbb{R}^{M \times C} $, where $M$ is the total number of instance queries. The instance queries should be with strong representation for future instance prediction. To this end, the instance queries are initially updated with six deformable DETR decoder~\cite{deformabledetr} layers which take $F_0$ as values.

Drawing from empirical observations, future instance segmentation heavily relies on the assistance of box regression at current frame. Thus we apply a class head and a 3D box head on updated instance queries $\mathrm{Q}_I$ and BEV feature map $F_0$ to output class prediction $\hat{c}$ and 3D box attributes $\hat{b}$ and these heads consist of a 3-layer feed-forward neural network, respectively.

For future instance prediction, the segmentation head takes the instance queries and BEV feature as input. To enrich the temporal cue of query and BEV feature at each timestamp, the updated instance query will be transformed with a 3-layer Multi-layer Perceptron(MLP) and the BEV feature map will be transformed with a context subnetwork composed of two residual CNN blocks. The parameters of all these modules are non-shared across frames, which is more adaptive and improves performance. The predicted masks are calculated by: 

\begin{equation}
\hat{m}_t^i = \mathrm{sigmoid}(\mathrm{Q}_{i,t} \cdot F_t')
\end{equation}

where $\hat{m}_t^i \in  \mathbb{R}^{1 \times H \times W}$ is the predicted mask logit, $\mathrm{Q}_{i,t}$ and $F_t'$ are the transformed i-th instance query and BEV feature at timestamp $t \in \{0, 1, ..., T_{out}\}$.

\subsection{Future Instance Matching}

Hungarian matching is an effective one-to-one matching strategy for end-to-end detection~\cite{detr, deformabledetr}, semantic segmentation~\cite{maskformer, genvis} and tracking~\cite{motr, trackformer}. For tasks aiming at predicting a sequence of masks at consecutive frames, independently applying Hungarian matching to each frame cannot ensure the temporal coherence. 

Motivated by this observation, we propose a novel future instance matching strategy to encourage a temporally consistent matching result for a specific instance across different future frames. In more detail, a specific object that appears in multiple frames should always be assigned to the same instance query. Taking a paired instance query and ground truth object, their matching cost includes two parts, \textit{i.e.}, the accumulated segmentation costs over $T_{out} + 1$ frames considering the predicted masks at the current and future frames and an auxiliary detection cost on the current frame. The matching cost can be organized as follows:

\begin{equation}
\begin{split}
    \mathcal{L}_{match} 
    = \mathcal{L}_{match}^{Det} + \mathcal{L}_{match}^{Seg},
\end{split}
\end{equation}

\begin{equation}
\begin{split}
    \mathcal{L}_{match}^{Det}
    = \mathcal{L}_{cls}(c, \hat{c}) + 
     \mathcal{L}_{box}(b, \hat{b}),
\end{split}
\end{equation}

\begin{equation}
    \mathcal{L}_{match}^{Seg}
    = \sum_{t=0}^{T_{out}}\mathcal{L}_{mask}(m_t, \hat{m_t}),
\end{equation}

where $c$ and $b$ represent the ground truth category and bounding box, while $\hat{c}$ and $\hat{b}$ denote the predicted ones. $m_t$ and $\hat{m}_t$ represent the ground truth mask and the predicted mask at a specific timestamp. For the detection cost, we use Focal Loss~\cite{focal} and L1 Loss for $\mathcal{L}_{cls}$ and $\mathcal{L}_{box}$ respectively,  while for the segmentation cost, we adopt Dice~\cite{dice} loss for $\mathcal{L}_{mask}$.

Following the common practice of previous end-to-end works~\cite{detr, maskformer, vistr}, the optimal assignment can be computed efficiently by the Hungarian algorithm~\cite{hungarian} which searches for a permutation $\hat{\sigma} $ with the lowest cost. 

We apply the same matching results among all the frames. In this way, an instance query is closely and consistently related to its assigned ground truth, which effectively improves temporal coherence.

\subsection{Overall loss function}

Given optimal matching, we use Hungarian matching loss for all matched pairs to train FipTR. Slightly different from the segmentation cost part in future instance matching, we incorporate L1 loss in $\mathcal{L}_{mask}$ for better convergence. Moreover, the flow-aware deformable attention needs additional supervision on backward flow and we utilize Smooth L1 loss for further optimization.

\begin{equation}
    \mathcal{L}_{FipTR} = \mathcal{L}_{match}  +  \mathcal{L}_{flow}(f, \hat{f})
\end{equation}

%----------------------------------- 
\section{Experiment}

\subsection{Datasets and settings}
We evaluate our approach on the NuScenes~\cite{nuscenes} dataset. Nuscenes contains 1000 scenes and each has 20 seconds annotated at 2HZ. For vision-centric methods, the provided sensory input includes six surrounding cameras with intrinsic and extrinsic calibration matrices and the ego-motions. We follow the training and evaluating settings used in previous works~\cite{fiery, beverse, stretchbev, powerbev} for fair comparison that the input is past one-second states plus current state and the output is future two seconds predicted states. The BEV coordinate is based on the ego-vehicle system. The global prediction range is $ 100m \times 100m $ and the size of the generated BEV grid map is $ 200 \times 200 $, each of which corresponds to $ 0.5m \times 0.5m $.  

For the model settings, we follow~\cite{beverse} which uses BEVDet4D~\cite{bevdet4d} as the temporal BEV encoder and the backbone is SwinTransformer~\cite{swin}. In addition, we also followed UniAD~\cite{uniad} which is based on BEVFormer~\cite{bevformer} to verify the effectiveness and robustness of FipTR with different temporal BEV encoders. The image size is set the same as the corresponding compared methods. We set sampled key numbers $K = 4$ and eight attention heads for four-layer flow-aware deformable attentions to predict future BEV feature sequences. We use six decoder layers of hidden dimension 256 and only the last layer is utilized to calculate the mask logits for computational saving.

For training, we use AdamW optimizer with an initial learning rate of 2e-4, weight decay of 0.01, and gradient clip of 35. For the learning schedule, we apply the cosine annealing scheduler. We train the model with a batch size of 8 on 8 NVIDIA GeForce RTX 3090 GPUs.

For inference, FipTR predicts the instance segmentation results for each frame simply using the result of the future masks decoder without any complex post-processing. Specifically, we multiply the object score and mask logits of a specific instance query as the final mask logits, $\hat{m}_i = c_i \cdot \hat{m}_i$  and set a threshold to filter which grid belongs to the foreground $\hat{m}_i > \delta_m$. When a grid is occupied with multiple instances, we choose the instance with max logit.

\subsection{Metrics}

Following~\cite{fiery, beverse}, we use IoU and VPQ(Future Video Panoptic Quality)~\cite{vpq} to evaluate the performance of future instance prediction. The IoU metric reflects the foreground segmentation performance while the VPQ metric measures the recognition and segmentation quality across frames for each predicted instance. The VPQ metric is computed as:

\begin{equation}
    \mathrm{VPQ} = \sum_{t=0}^{T_{out}}\frac{\textstyle \sum_{(p_t,q_t) \in TP_t}IoU(p_t, q_t)}{|TP_t| + \frac{1}{2}|FP_t| + \frac{1}{2}|FN_t|}
\end{equation}

where $TP_t$, $FP_t$ and $FN_t$ represent the set of true positives, false positives and false negatives at timestamp t. All experiments will be conducted under two ROI settings, namely 30mx30m(n) and 100mx100m(f) around the ego vehicle.

\begin{table*}[htbp]
\caption{\textbf{Future Instance Prediction results on nuScenes validation set.} IoU is used for future semantic segmentation and VPQ for future instance prediction. Results are reported under two roi settings: n for $30m \times 30m$ and f for $100m \times 100m$. $\dag$: Results trained with CBGS. $\S$: Reproduced results. \textcolor{blue}{+} and \textcolor{red}{-} represent the performance gap between FipTR and the corresponding model with the same backbone and BEV encoder setting.}
\label{tab:main_result}
\centering
\begin{tabular}{c|c|c|cc|cc}
\toprule
 Method                   & Backbone & BEV Encoder    & IoU(n) & IoU(f) & VPQ(n) & VPQ(f) \\ 
 \midrule
 FIERY                    & EfficientNet &BEVDet4D & 59.4 & 36.7 & 50.2  & 29.9  \\
 StretchBEV               & EfficientNet &BEVDet4D& 55.5  & 37.1  & 46.0 & 29.0 \\
 PowerBEV                 & EfficientNet &BEVDet4D& \textbf{62.5} & 39.3 & 55.5 & 33.8 \\
$\text{BEVerse-T}^\S$     & Swin-T       &BEVDet4D& 57.7 & 35.9 & 48.8 & 29.7 \\
$\text{BEVerse-S}^\S$     & Swin-S       &BEVDet4D& 60.1 & 40.3 & 51.7 & 33.7 \\
$\text{BEVerse-T}^\dag$   & Swin-T       &BEVDet4D& 60.3 & 38.7 & 52.2 & 33.3 \\
$\text{BEVerse-S}^\dag$   & Swin-S       &BEVDet4D& 61.4 & 40.9 & 54.3 & 36.1 \\

$\text{FipTR-T}$          & Swin-T       &BEVDet4D& 59.0(\textcolor{blue}{+1.3})  & 36.6(\textcolor{blue}{+0.7})  & 53.3(\textcolor{blue}{+4.5})  & 33.6(\textcolor{blue}{+3.9})  \\
$\text{FipTR-S}$          & Swin-S       &BEVDet4D& 61.7(\textcolor{blue}{+1.6}) & 40.4(\textcolor{blue}{+0.1}) & 56.1(\textcolor{blue}{+4.4})  & 37.0(\textcolor{blue}{+3.3}) \\
$\text{FipTR-T}^\dag$     & Swin-T       &BEVDet4D& 60.7(\textcolor{blue}{+0.4}) & 38.3({-0.4}) & 55.4(\textcolor{blue}{+3.2}) & 36.1(\textcolor{blue}{+2.8}) \\
$\text{FipTR-S}^\dag$     & Swin-S       &BEVDet4D& \makebox[0.035\textwidth][c]{62.2(\textcolor{blue}{+0.8})} & \makebox[0.035\textwidth][c]{\textbf{41.1}(\textcolor{blue}{+0.2})} & \makebox[0.035\textwidth][c]{\textbf{58.0}(\textcolor{blue}{+3.7})} & \makebox[0.035\textwidth][c]{\textbf{39.5}(\textcolor{blue}{+3.4})} \\

 \midrule
UniAD        & R101        &BEVFormer& \textbf{63.4} & 40.2 & 54.7 & 33.5 \\
FipTR-R50    & R50         &BEVFormer& 57.6 & 33.6 & 50.1 & 30.6 \\
FipTR-R101   & R101        &BEVFormer& 62.9({-0.5}) & \makebox[0.035\textwidth][c]{\textbf{40.7}(\textcolor{blue}{+0.5})} & \makebox[0.035\textwidth][c]{\textbf{57.2}(\textcolor{blue}{+2.5})} & 
\makebox[0.035\textwidth][c]{\textbf{37.8}(\textcolor{blue}{+4.3})} \\
\bottomrule

\end{tabular}
\end{table*}

\subsection{Main Results}

We first compare the overall performance with multiple excellent works including FIERY~\cite{fiery}, StretchBEV~\cite{stretchbev}, PowerBEV~\cite{powerbev}, BEVerse~\cite{beverse}, and UniAD~\cite{uniad} in \cref{tab:main_result}. We can see that i) FipTR achieves SOTA on all metrics with BEVDet4D as BEV encoder, and gains a significant improvement on VPQ metrics, indicating superior segmentation prediction capability and frame consistency without any auxiliary output and post-processing. ii) FipTR can achieve effective performance based on various temporal BEV encoders robustly that FipTR with BEVFormer achieves SOTA on VPQ metrics while maintaining competitive performance on IoU metrics. 

Specifically, without CBGS, FipTR shows remarkable improvement over BEVerse on both metrics with different image backbones. Considering data augmentation, FipTR also achieves a significant improvement on the VPQ metric. Among all the models utilizing BEVDet4D, FipTR surpasses others by a large margin, achieving state-of-the-art performance.

Whereas previous works are only based on a specific temporal BEV encoder and the robustness of different BEV perception paradigms is unidentified, we follow the setting of UniAD that takes BEVFormer as the BEV encoder and adapt it in FipTR while retaining the design of BEV predictor and mask decoder. The results reveal that FipTR with BEVFormer is effective as well. Notably, FipTR exhibits a great enhancement of VPQ and sets the state-of-the-art, and maintains a comparable IoU performance.
%----------------------------------- Main results

%----------------------------- ablation
\subsection{Ablation Studies}

\textbf{Flow-aware Deformable Attention. } The flow-aware deformable attention (FADA) proposed in the BEV predictor aims to provide more precise and interpretable offsets in the attention mechanism. ~\cref{tab:future_prediction_module} compares FADA with vanilla deformable attention(DA) and RNN-based~\cite{beverse}, CNN-based~\cite{powerbev} methods in BEV feature predict module and we keep all the settings the same except for the prediction module. The result illustrates that deformable attention already possesses a commendable future BEV prediction capability considering CNN methods have a multi-scale structure and RNN methods require complex distribution modeling. In comparison to vanilla deformable attention, FADA indicates further improvement in the short range on both metrics benefiting from the guidance of predicted flow. Furthermore, we demonstrate the effective validation of FADA on different image backbones and BEV encoders in \cref{tab:flow-aware} that illustrate FADA brings the similar gains with different temporal BEV encoders.

\begin{figure}[h]
  \centering
   \includegraphics[width=1.0\linewidth]{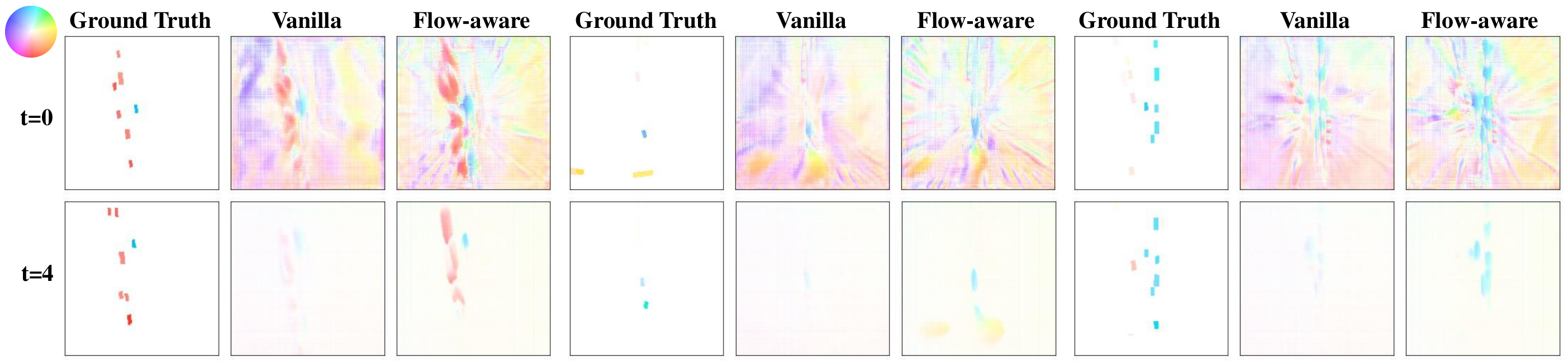}

   \caption{Visualization of ground truth backward flow and mean of sampling offsets at different timestamps. The color represents the direction and the transparency represents the magnitude.}
   \label{fig:flow}
\end{figure}

To elucidate the interpretability of FADA, we visualize some examples including the mean of sampling offsets in deformable attention and the backward flow ground truth at different frames in \cref{fig:flow}. In alignment with our prior conjecture, the role played by the sampling offsets can be approximately regarded as the backward flow in deformable attention. Vanilla deformable attention conducts reasonable sampling at t=0 that the direction and the magnitude are close to ground truth flow, while it becomes less motion-aware at t=4. On the contrary, the offset induced by FADA embodies two distinctive traits: i) amplified long-term predictive abilities for foreground objects, ii) diminished offset magnitude in the background locations.

\begin{table}[t]
\caption{Different future BEV prediction module. RNN and CNN methods are modified to accommodate FipTR.}
\label{tab:future_prediction_module}
\centering
\begin{tabular}{c|cc|cc}
\toprule
Future module & IoU(n) &  IoU(f)&  VPQ(n) &  VPQ(f) \\ 
  \midrule
\makecell{RNN~\cite{beverse}}   & 51.8  & 30.2   & 46.7   & 27.7 \\
 \makecell{CNN~\cite{powerbev}}   & 58.8   & 36.8   & 52.4   & 33.0 \\
 \makecell{DA}   & 57.8   & 36.4   & 52.5   & 33.3 \\
 \makecell{FADA} & 59.0   & 36.6   & 53.3   & 33.6 \\
\bottomrule
\end{tabular}
\end{table}

\begin{table}[t]
\caption{The performance of the vanilla and flow-aware deformable attention with different temporal BEV encoders. FipTR-S is with BEVDet4D and FipTR-R101 is with BEVFormer.}
\label{tab:flow-aware}
\centering
\begin{tabular}{c|c|cc|cc}
\toprule
{Model} & {FADA} & {IoU(n)} &  {IoU(f)}&  {VPQ(n)} &  {VPQ(f)} \\ 
  \midrule
\multirow{2}{*}{FipTR-S} &  & 60.9   & 40.4   & 55.8   & 36.9 \\
     & \checkmark & 61.7   & 40.4   & 56.1   & 37.0 \\
\multirow{2}{*}{FipTR-R101} &  & 62.0   & 40.8   & 56.6   & 37.6 \\
     & \checkmark & 62.9   & 40.7   & 57.2   & 37.8 \\
\bottomrule
\end{tabular}

\end{table}

\noindent\textbf{Module Weight Shared.} We then demonstrate the performance gains of three non-weight-shared modules: the Flow CNN, Context CNN, and Instance Query in \cref{tab:time}. The use of non-weight-shared subnetworks for Flow CNN and Context CNN across frames is vital since the unified future BEV queries play a similar role in each future frame and such a non-weight-shared setting can bring more temporal cues and adaptivity. Moreover, the time-specific instance queries further improve the VPQ metrics considering different potential locations or motion states at different timestamps. 

\begin{table}[t]
\caption{Specific module is modified to weight-shared across frames, $-$ indicates all modules are non-weight-shared.}
\label{tab:time}
\centering
\begin{tabular}{c|cc|cc}
\toprule
 Shared Module              & {IoU(n)} &  {IoU(f)}&  {VPQ(n)} &  {VPQ(f)} \\ 
 \midrule
 \makecell{Context CNN  }      & 48.7 & 30.6 & 41.9 & 26.0 \\
 \makecell{Flow CNN }          & 57.1 & 26.2 & 51.9 & 30.2 \\
 \makecell{Instance Query}         & 58.4 & 36.2 & 52.5 & 32.6 \\
 -                        & 59.0 & 36.6 & 53.3 & 33.6 \\
\bottomrule
\end{tabular}

\end{table}

%-------------------  matching
\noindent\textbf{Matching Cost.} \cref{tab:matchin_cost_loss} shows the performance related to mask cost. Without the mask cost, FipTR has an obvious performance degeneration only considering the box cost, underscoring the necessity of incorporating the mask cost for BEV instance segmentation. Moreover, when merely considering mask matching of single-frame, while the IoU upholds a stable level, there is a significant drop in the VPQ metrics, indicating the proposed future instance matching strategy can efficaciously augment the temporal consistency across frames of each instance.

\begin{table}[t]
\caption{Different matching cost strategy. single-mask matching indicates only mask cost at $t = 0$ is incorporated. }
\label{tab:matchin_cost_loss}
\centering
\begin{tabular}{c|cc|cc}
\toprule
 Cost                        & {IoU(n)} &  {IoU(f)}&  {VPQ(n)} &  {VPQ(f)} \\ 
 \midrule
 w/o mask matching           & 52.6 & 30.2 & 48.2 & 28.8 \\
single-mask matching     & 58.2 & 36.3 & 51.1 & 31.7 \\
 multi-mask matching       & 59.0 & 36.6 & 53.3 & 33.6 \\
\bottomrule
\end{tabular}

\end{table}

\noindent\textbf{Box and Mask Branch.} Considering the heated discussion of the interrelationship between the box branch and mask branch~\cite{beverse, sceneocc}, we compare the detection and segmentation performance when removing the opposing branch (both the loss and  cost) in \cref{tab:matchin_loss}. The auxiliary box branch plays an important role and directly predicting BEV masks in complex environments is difficult and the segmentation capability will be invalid. Moreover, we show the mean Average Precision (mAP) and mean Average Velocity Error (mAVE) which are 3D object detection metrics. Benefiting from the multi-task setting in matching cost and loss function, FipTR can conduct 3D object detection and future instance prediction simultaneously. The results indicate that the segmentation prediction is not negative to detection performance, and excitingly, the mAVE metric has a significant improvement due to the motion modeling in FipTR.

\begin{table}[t]
\caption{Different Hungarian loss strategy. w/o box branch represents removing 3D box regression cost and loss, and w/o mask branch represents removing multi-frame mask cost and loss. }

\label{tab:matchin_loss}
\centering
\begin{tabular}{c|c|c|c|c}
\toprule
 Branch          & mAP & mAVE & IoU & VPQ \\ 
 \midrule
 w/o box branch  & - & - & 5.2 & 1.1 \\
 w/o mask branch & 24.6 & 0.78   & - & - \\
 box \& mask branch  &  24.4 & 0.51 & 59.0 & 33.6 \\
\bottomrule
\end{tabular}

\end{table}

\subsection{Qualitative Results}
\begin{figure*}[h]
  \centering
   \includegraphics[width=1.0\linewidth]{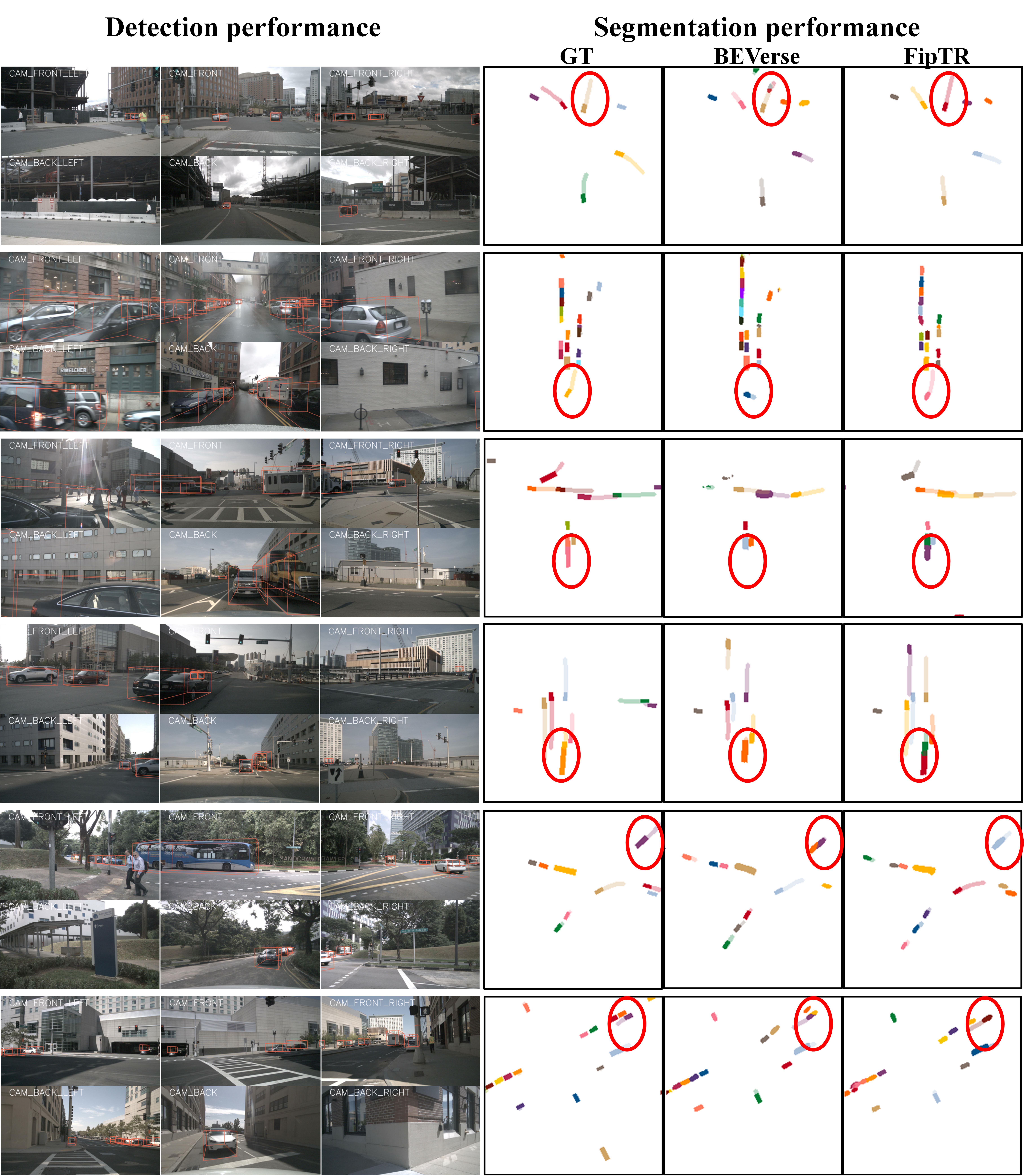}

   \caption{Demonstration of performance of FipTR compared with BEVerse and Ground Truth. The left part is the detection performance and only vehicles are considered. The right part is the segmentation performance where darker regions are the result of the current frame while the lighter ones are the prediction in the future frames. }
   \label{fig:quality}
\end{figure*}

Figure~\ref{fig:quality} shows the visualization results. BEVerse and FipTR are both based on Swin-tiny backbone with CBGS. For different cases, we highlight the interesting instances for more attention. We can see that i) FipTR and BEVerse both can achieve accurate predictions for most instances. ii) FipTR possesses the capability of executing 3D detection and BEV segmentation simultaneously, dependent on the design of Hungarian matching cost and loss. iii) FipTR excels at modeling the motion states, especially for objects that are cornering or at a far distance. iv) FipTR ensures superior temporal and instance consistency thereby avoiding the allocation of multiple IDs to the same instance.

\section{Conclusion}

In this work, we have proposed FipTR to perform BEV future instance prediction. FipTR can effectively generate high-quality instance masks without any auxiliary post-processing in an end-to-end manner. In flow-aware BEV predictor, the proposed FADA module provides precise and interpretable sampling offsets for BEV prediction. Taking the future BEV feature sequence and instance queries as input, the future masks decoder simply generates the instance segmentation masks in parallel. Considering multi-frame mask cost, future instance matching further provides a temporally coherent matching result. Extensive experiments have shown that FipTR is effective based on different temporal BEV encoders.

\clearpage
\bibliographystyle{splncs04}
\bibliography{main}
\end{document}